\documentclass{article}

\usepackage{PRIMEarxiv}

\usepackage[utf8]{inputenc} 
\usepackage[T1]{fontenc}    
\usepackage{hyperref}       
\usepackage{url}            
\usepackage{booktabs}       
\usepackage{amsfonts}       
\usepackage{nicefrac}       
\usepackage{microtype}      
\usepackage{lipsum}
\usepackage{fancyhdr}       
\usepackage{graphicx}       
\graphicspath{{media/}}     

\usepackage{times}
\usepackage{epsfig}
\usepackage{graphicx}
\usepackage{amsmath}
\usepackage{amssymb}
\usepackage{caption}
\usepackage{subcaption}
\usepackage{pifont}
\newcommand{\xmark}{\ding{55}}%

\title{Multi-Domain Norm-referenced Encoding Enables Data Efficient Transfer Learning of Facial Expression Recognition
\thanks{\textit{\underline{Citation}}: 
\textbf{Authors. Title. Pages.... DOI:000000/11111.}} 
}

\author{
  Michael Stettler, Alexander Lappe \\
  Department of Cognitive Neurology \& International Max Planck research School for Intelligent Systems (IMPRS-IS) \\
  Eberhard Karls University of T\"ubingen \\
  T\"ubingen, Germany\\
  \texttt{\{michael.stettler, alexander.lappe\}@uni-tuebingen.de} \\
   \And
  Nick Taubert, Martin Giese \\
  Department of Cognitive Neurology \\
  Eberhard Karls University of T\"ubingen \\
  T\"ubingen, Germany \\
  \texttt{\{nick.taubert, martin.giese\}@uni-tuebingen.de} \\
}

\begin{document}
\maketitle

\begin{abstract}
People can innately recognize human facial expressions in unnatural forms, such as when depicted on the unusual faces drawn in cartoons or when applied to an animal's features. However, current machine learning algorithms struggle with out-of-domain transfer in facial expression recognition (FER). We propose a biologically-inspired mechanism for such transfer learning, which is based on norm-referenced encoding, where patterns are encoded in terms of difference vectors relative to a domain-specific reference vector. By incorporating domain-specific reference frames, we demonstrate high data efficiency in transfer learning across multiple domains. Our proposed architecture provides an explanation for how the human brain might innately recognize facial expressions on varying head shapes (humans, monkeys, and cartoon avatars) without extensive training. Norm-referenced encoding also allows the intensity of the expression to be read out directly from neural unit activity, similar to face-selective neurons in the brain. Our model achieves a classification accuracy of 92.15\% on the FERG dataset with extreme data efficiency. We train our proposed mechanism with only 12 images, including a single image of each class (facial expression) and one image per domain (avatar). In comparison, the authors of the FERG dataset achieved a classification accuracy of 89.02\% with their FaceExpr model, which was trained on 43,000 images.
\end{abstract}


\section{Introduction}
\label{sec:intro}

Our visual system has a remarkable property: We can innately recognize facial expressions on non-human faces, such as those of cartoon characters. Facial expression recognition (FER) is critical for human social interactions \cite{wilson2012social}. 

\begin{figure}[t]
	\centering
	\includegraphics[width=1.0\linewidth]{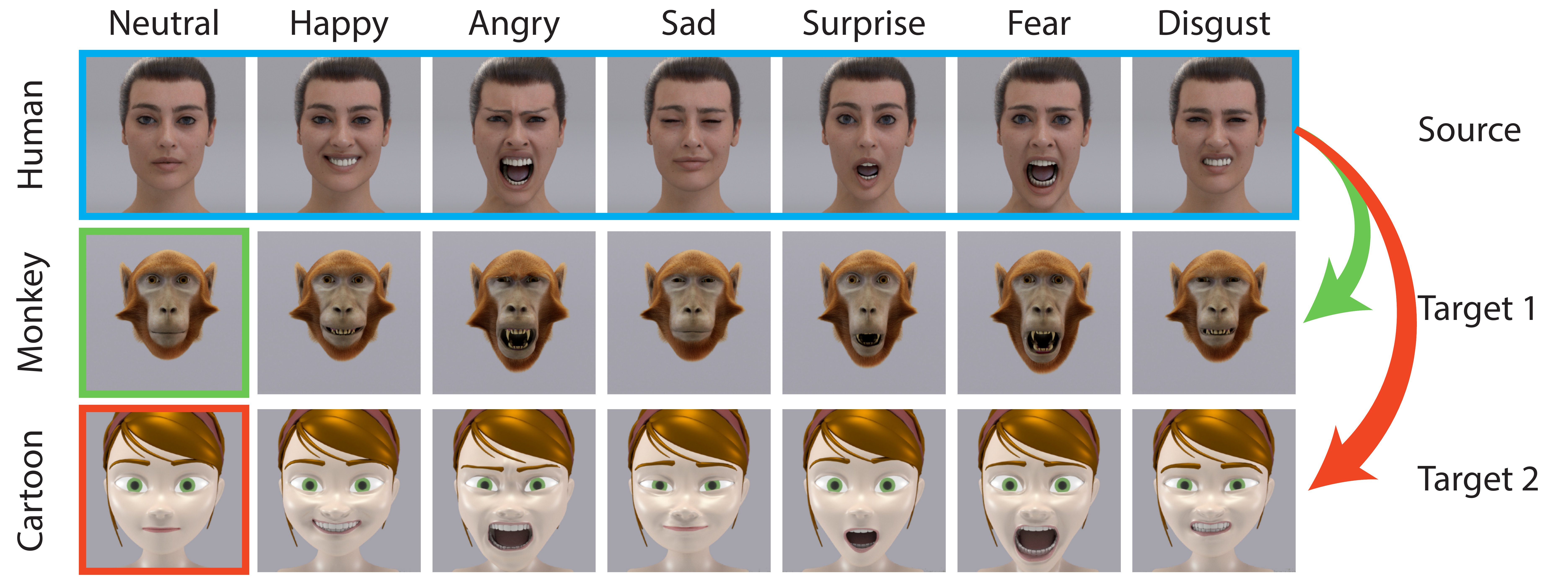}
	\caption{Examples of the portraits that form our basic face shape (BFS) dataset showing seven different expressions on the three basic head shapes-what we refer to as our source and target domains. Our task involves the classification of facial expressions across unseen target domains from a source domain (outlined in blue) using only a single image from the target domain as a reference (outlined in red and green).}
	\label{fig:BFS_example}
\end{figure}

While there have been few studies specifically investigating cross-species facial perception \cite{zhu2013dissimilar, taubert2021shape}, there exists a significant body of work on facial perception (\textit{e.g.}, \cite{ekman1976measuring, jack2016four}) within a variety of domains, including caricature drawings \cite{rhodes1987identification, gao2003facial}, emoticons/emojis \cite{kaye2017emojis}, cartoon characters \cite{zhao2019event, zhang2021influence}, as well as studies on the phenomenon of pareidolia (the perception of faces on arbitrary objects)\cite{wardle2020rapid}. These studies indicate that human face perception generalizes strongly even to previously unencountered, non-human head shapes. Figure \ref{fig:BFS_example} illustrates our proposed task of transferring facial expression recognition from one source domain (\textit{e.g.}, human head shape) to other target domains (\textit{e.g.}, monkey, cartoon) using the least possible amount of images.

Despite the recent advances in machine learning, tested state-of-the-art FER models do not generalize (transfer) to, \textit{e.g.}, cartoon faces (see results on our developed BFS dataset \ref{ss:transferLearning}). There are several possible explanation for these results. CNN architectures are know to have strong bias towards textures \cite{geirhos2018generalisation, baker2020local}. We argue that CNNs rely too heavily on local features which outweigh global information to achieve our task. More recent computer vision models, such as vision transformers (ViT) \cite{dosovitskiy2020image}, partially solve the local issue able to access any part of an image at any given layer. Nonetheless, they require large amounts of data to learn the spatial embedding.

In this work, we seek to explore a mechanism well-known in neuroscience: norm-referenced encoding (NRE) \cite{rhodes2011adaptive, leopold2006norm}. In this approach, facial identities are encoded in terms of differences \textit{tuning vectors} relative to an average-norm-face (Figure \ref{fig:norm_example}A). We aim at demonstrating its benefits in computer vision. By extending the encoding to multiple domains and developing a machine-learning model exploiting this encoding, we make a step forward towards an end-to-end framework. The implicit bias of this encoding enables our model to transfer knowledge with high data efficiency. Our model generalizes to new domains using a single image. 

\textbf{Contributions.} \textbf{(1)} We extend the biologically-plausible Norm-referenced encoding to multiple domains. \textbf{(2)} We develop an architecture to exploit the multi-domain Norm-referenced encoding as a step towards an end-to-end framework.\textbf{(3)} We provide a dataset to study transfer learning with corresponding facial expressions using fundamentally different head shapes and different levels of expression intensity. \textbf{(4)} We test our model on a publicly available dataset and illustrate the high data efficiency, robust transfer learning capability, and interpretability of the proposed method.

\section{Background}
\begin{figure}[t]
	\centering
	\includegraphics[width=.6\linewidth]{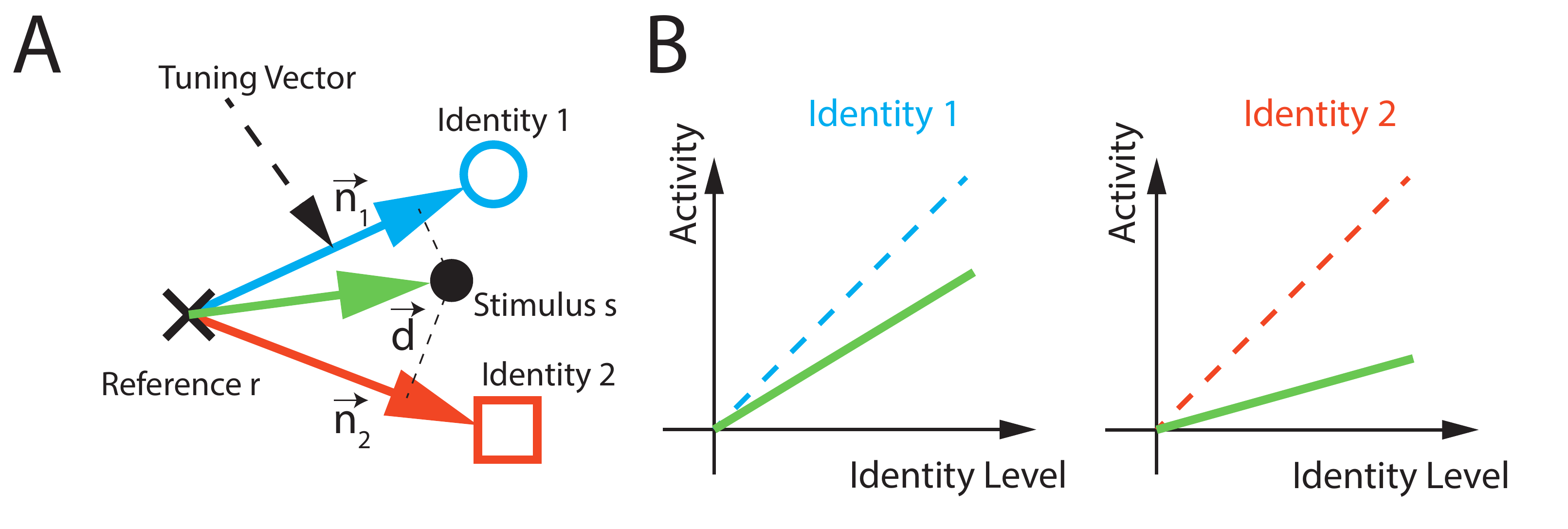}
	\caption{\textbf{A)} A schematic representation of norm-based encoding. The sketch displays two identities in feature space and their respective tuning vectors $\bf n_1$ and $\bf n_2$. The stimulus $\bf s$ is encoded by its position relative to the reference $\bf r$ through the difference vector $\bf d = \bf s - \bf r$. \textbf{B)} Read-out activity $v_i$ for the input stimulus $\bf s$ for the two identities. The activity is given by $v_i = \bf d^T \bf n_i$ and thus grows linearly in the identity level $\Vert \bf d \Vert$. The slope is given by $\bf d^T \bf n_i/\Vert \bf d \Vert$, meaning $\bf d = \bf n_i$ yields the identity function.}
	\label{fig:norm_example}
\end{figure}

Norm-referenced encoding (NRE) is a classical principle in neuroscience for face identity representation. The key idea of NRE is that the face identity is encoded by a difference vector relative to a norm stimulus, typically the average face. The direction of this difference encodes facial identity, while its length represents the distinctiveness of the face relative to the average face (Figure \ref{fig:norm_example}A).

Electrophysiological evidence supports this encoding principle \cite{leopold2006norm, koyano2021dynamic} (related neural model have been proposed \cite{giese2005physiologically}) and recordings show that the tuning relative to the norm face emerges more slowly than the absolute encoding, about 100 ms after stimulus presentation \cite{koyano2021dynamic}. These results suggest a fast process that exploits encoding in absolute space \cite{chang2017code} and a slower one that realizes an encoding relative to the average face, potentially mediated by recurrent feedback \cite{freiwald2021neuroscience}.

Let $\bf s$ be a vector that represents a face stimulus in an appropriate feature space. Then the difference vector is defined by $\bf d = \bf s - \bf r$, where $\bf r$ is a norm or reference vector—classically, the \textit{average face} computed by averaging the feature vectors of a large number of faces. 

The response of a face-encoding neuron is computed as the scalar product of the difference vector $\bf d$ of the actual stimulus and a unit vector $\bf n_i$ (tuning vector) that determines the preference of the neuron in terms of face identity:

\begin{equation}
\label{eq:norm_ref_dot_product}
  v_i = {\bf d}^T {\bf n}_{i}.
\end{equation}

The neuron would respond maximally to face stimuli for which the associated difference vector $\bf d$ is collinear with the vector $\bf n_i$. The length of the difference vector $\Vert \bf d \Vert$ encodes how characteristic the face is for a specific facial identity (called 'identity level' in psychophysics) (See Fig. \ref{fig:norm_example}B). The same encoding principle can be transferred to facial expression recognition \cite{stettler2020physiologically}. Here, the norm stimulus is the neutral expression, the vectors $\bf n_i$ encode the expression types, and the length of the difference vector represents the expression strength of the face.

\section{Multi-Domain Norm-Referenced Encoding}
\label{sec:norm-ref_encoding}
\begin{figure}[h]
    \centering
    \includegraphics[width=.6\linewidth]{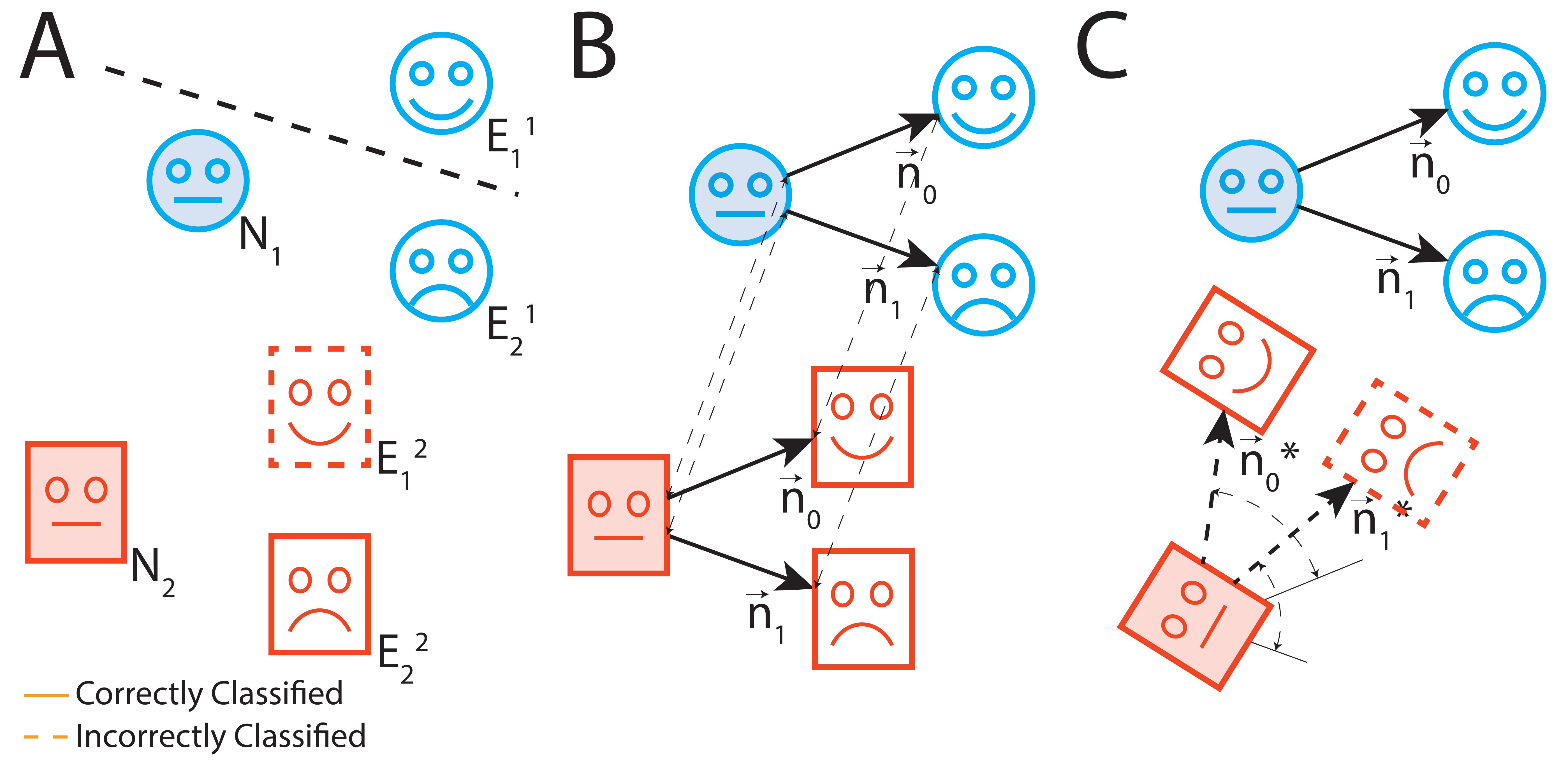}
	\caption{Schematic representation of the classification of two expressions ($E_1$ and $E_2$) in a 2D face space, presented on two different basic head shapes. $N_1$ and $N_2$ represent the "norm faces" (neutral expressions) relative to which the individual expressions are encoded. \textbf{A)} A linear classifier class boundary indicated by the dashed line results in the correct classification of the expressions of the first head shape ($N_1$), but fails on the second head shape ($N_2$). \textbf{B)} A norm-referenced classifier, which transfers the tuning vectors $\bf n_1$ and $\bf n_2$ from the first head shape $N_1$ to the second head shape $N_2$ accomplishes correct classification without retraining the classifying neurons. \textbf{C)} An example of a norm-referenced classifier in a poorly-chosen feature space, resulting in misclassification for the second head shape.}
	\label{fig:fig1}
\end{figure}
NRE is thus an intuitive and powerful model to encode faces and facial expressions. Here, we propose to extend the model to multiple reference frames, referring to it as \textit{multi-domain norm-referenced encoding} (MD-NRE). We hypothesize that utilizing multiple norm-references might be a highly data-efficient transfer-learning approach for multi-domain FER.

Consider the problem of classifying facial expressions $E$ that correspond to feature vectors $\bf s$. Let $N$ denote the domain (head shape) from which $\bf s$ was drawn. For each domain, we consider a domain-specific reference vector $\bf r_N$. Finding $\bf r_N$ is part of the learning procedure. The MD-NRE assumes that the distribution of $E$ only depends on the domain $N$ through the difference vector between the feature vector $\bf s$ and the domain-specific reference. More specifically,
\begin{equation}
\begin{split}
    p(E|\mathbf{s}) &= \sum_N p(E|\mathbf{s},N) p(N|\mathbf{s}) \\
    &= \sum_N p(E|\mathbf{s} - \mathbf{r_N})p(N|\mathbf{s}).
\end{split}
\end{equation}
This assumption breaks the learning process down into two parts: learning to infer the domain from the feature vector, and learning to infer the expression from the difference vector $\bf d=s-r_N$. This decomposition allows us to learn the distribution of $E$ for all domains while only training on data from one domain.

Fig. \ref{fig:fig1} illustrates the situation. In a two-dimensional feature space, we aim to classify the expressions $E_1$ and $E_2$ for two different head shapes. The neutral expressions for the two head shapes are denoted as $N_1$ and $N_2$. Panel A shows the situation using a linear classifier that separates the two expressions correctly for only one head shape. Suppose the effect of changing the basic head shape is a collinear translation of all feature vectors in the feature space. Even if the translation vector is the same for all tested expressions, the classifier fails to classify the two expressions correctly for the second head shape. However, this is different when MD-NRE is applied. As illustrated in panel B, assume that the directions of the difference vectors between the expressions ($E_1^1$, $E_2^1$ and $E_1^2$, $E_2^2$) and the corresponding neutral reference face vectors ($N_1$ or $N_2$) are the same for the two head shapes. Then MD-NRE, which uses the same tuning vectors $\bf n_1$ and $\bf n_2$ for the two head shapes, would show perfect transfer of expression recognition from the first head shape to the second. Therefore, training the system on the first head shape and knowing the feature vector of the second neutral (norm) face ($N_2$) allows a correct classification of the expressions on the second head shape without the need for explicit training of all expressions for the second head shape. Therefore, if done correctly, the MD-NRE promises great data efficiency in generalizing to new domains. In addition, the length of the difference vectors covaries directly with the expression strength, which is useful for many technical applications that exploit expressiveness.

Of course, the previous assumption that the tuning vectors $\bf n_0$ and $\bf n_1$ remain similar for different head shapes is far from trivial to achieve. It depends critically on the embedding of images in the 2D feature space. Panel C shows a hypothetical example where the tuning vectors change direction. Here the expressions for the first head shape do not help the classification of the expressions for the second. Such misalignment of the corresponding tuning vectors will be particularly apparent if the embedding feature space contains many dimensions unrelated to the expression classification task. Using, for example, texture-sensitive features, one would expect strong and unsystematic changes in the tuning vector moving from a human to a non-human head, which likely outweighs the expression-specific changes.

To overcome this challenge, we propose to add three mechanisms to NRE. 1) Use multiple domain-specific reference vectors. 2) Treat each reference vector as an updatable set of parameters. 3) Construct the relative encoding with a two-stream pipeline. Table \ref{tab:main_diff_NRE} summarizes the key aspects of our proposed model. A strong assumption, and limitation, of the model is to assume that the reference vectors for each domain are given. In other words, it is assumed that the model observes one image from each domain during training.

\begin{table}[t]
    \begin{center}
    \begin{tabular}{l r r r}
        Methods             & Encoding      & Multi-dom.    & Eff. transfer \\ 
        \hline
        Lin. Classifier     & absolute      & \checkmark    & \xmark \\
        NRE                 & relative      & \xmark        & \xmark \\  
        MD-NRE              & relative      & \checkmark    & \checkmark \\
        \hline
    \end{tabular}
    \end{center}
    \caption{Summary of the key differences between the classical linear classifier, the previous norm-referenced encoding model, and the current one. Multi-Dom = Multi-domain, Eff. Transfer = Efficient transfer.}
    \label{tab:main_diff_NRE}
\end{table}

\section{Related Work}

\textbf{Facial Expression Recognition:} 
There are numerous publicly available human facial expression recognition (FER) datasets in computer vision (\textit{e.g.} AffectNet \cite{mollahosseini2017affectnet}, FER2013 \cite{courville2013fer}, CK+ \cite{lucey2010extended}, KDEF \cite{calvo2008facial}) as well as datasets of cartoon characters (FERG dataset \cite{aneja2016modeling}). FER studies typically aim to reach the best classification on one or several of these datasets \cite{ionescu2013local, georgescu2019local, giannopoulos2018deep, niu2021facial, ashir2020facial}. Others investigate transfer learning (domain adaptation) \cite{zhao2018transfer, wang2018unsupervised} from a source (human) to a target dataset (another human). To the best of our knowledge, only one study experiments with transfer learning in a cross-species FER experiment between humans (JAFFE, KDEF) and cartoon faces (FERG) \cite{zhao2018transfer}. However, this method did not investigate data efficiency and employs an alternative method, learning a mapping function from the source and target datasets to a common feature space. Constructing such a mapping function bears greater resemblance to the task of facial retargeting \cite{ribera2017facial,chaudhuri2019joint}. Facial retargeting aims at learning a mapping function which links the same facial expressions across different head shapes (domains) in the context of facial animation. As such, data-efficient, cross-species FER generalization is not well studied.

\textbf{Data Efficient Transfer Learning:}
Developing machine learning models that transfer knowledge is a critical goal in the field. There are many generalization-related research topics such as transfer learning, zero-shot learning, meta-learning, and causal representational learning. In our study, we define our task as falling between domain adaptation (DA) \cite{kouw2019review} and domain generalization (DG) \cite{wang2022generalizing}. The goal of domain generalization is a good performance on novel domains by models trained on one or several distinct, but related domains. As in DG tasks, we assume that our training and testing sets are not identically and independently distributed (\textit{i.i.d.}).

A key difference between our approach and DG is that, as DA, we do not consider the target domain to be entirely unknown. Instead, we allow our model to see one training example per target domain. We achieve high data efficiency by introducing domain-specific inductive biases, which we refer to as our reference vectors. As the reference vectors are updatable parameters of our model, our approach falls under the \textit{algorithm strategy} in few-shot learning \cite{wang2020generalizing}, which means that existing parameters are refined during domain transfer.

\textbf{Morphable Models:} 
Early computational vision work proposed encoding shapes in vector spaces relative to a reference or norm pattern \cite{vetter1995separate, vetter1997bootstrapping}. The construction of vector spaces that parameterize properties of image classes is standard in computer graphics, particularly for faces \cite{blanz1999morphable, li2017learning, egger20203d}. 3D morphable models (3DMM) are powerful tools to construct, animate, and transfer facial expressions between avatars. 3DMMs learn to separate shape from appearance variation and construct a low-dimensional prior of a basic head shape. As such, they are intuitive and help in a natural way to disentangle shape changes that carry semantically critical information, such as expression or identity from the basic shape of the head. 


\section{Model Architecture}
\label{sec:model}
To exploit the multi-domain norm-referenced encoding (MD-NRE), the developed model must fit several constraints. First, it needs to construct representations that preserve the tuning vectors across domains. Once the reference and tuning vectors are trained, we need to update the specific reference vectors to encode the inputs relative to it and project the resulting difference vectors onto the tuning directions. Therefore, an architecture with two streams is required: one that updates the reference vector, and one that computes the difference vectors and their projection onto the tuning vectors $\bf n_i$. 

A straightforward method for a FER task would be to use pre-trained facial landmark detector methods. Tracking the landmarks' displacement for facial expression features would directly output a 2D representation such that the displacement is invariant to face shapes and textures. Such representations support generalization for frontal to moderately rotated views. However, tested state-of-the-art facial landmark detectors \cite{bulat2017far, kartynnik2019real} yield poor results on non-human face (see supplementary materials for more details on the tested landmark estimation models). Moreover, a \textit{simple} facial landmark detector would not be sufficient to exploit MD-NRE. In order to compute the outputs of the encoding neurons, the norm-referenced neurons need two inputs: one is the reference vector $\bf r_N$, and the other one is the learned preferred tuning direction $\bf n$. To transfer between expressions on different head shapes, we need a second stream that determines the type of the head and the associated reference vector. 

\begin{figure*}
  \centering
	\includegraphics[width=1.0\linewidth]{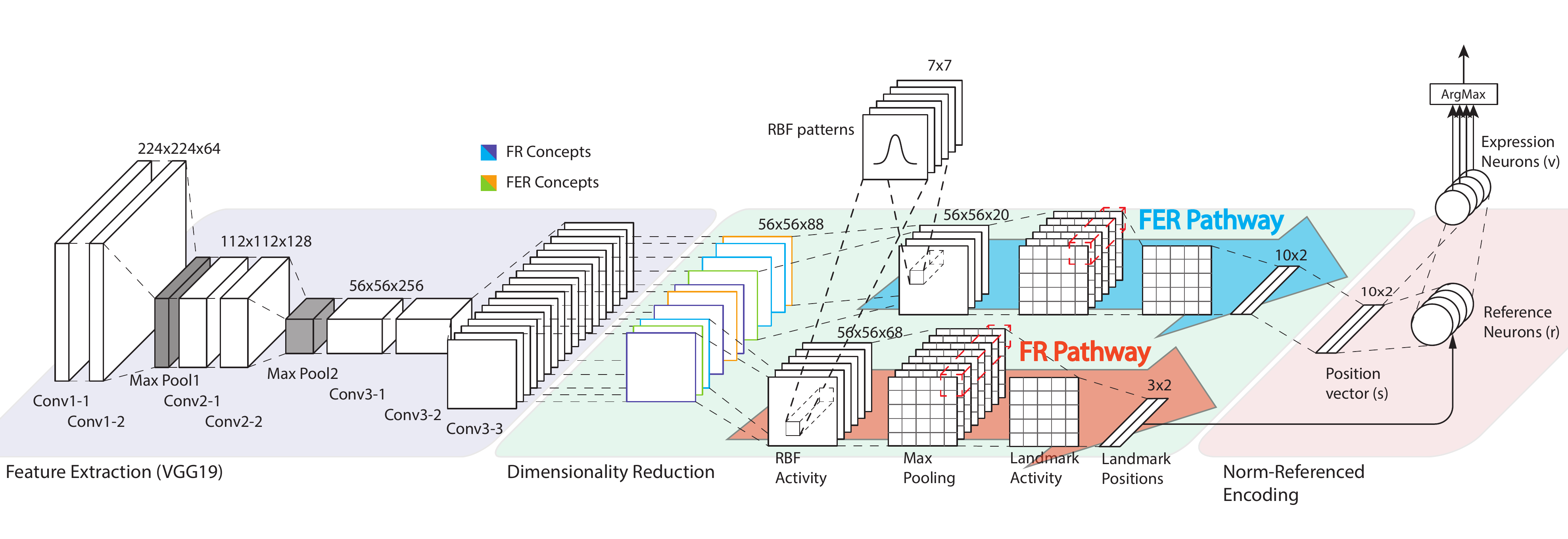}
	\caption{Model architecture comprising three main components: a CNN for generic feature extraction, a module for feature reduction and the construction of robust features, and a read-out network / classifier module. The feature reduction transforms the latent space of the VGG19 model into a sparse set of facial landmarks. The dimensionality reduction module comprises two pathways. The FR (facial recognition) pathway acts as a reference vector selector and updates the reference parameters with the face position and scale. The FER (facial expression recognition) pathway outputs the absolute position of facial expression landmarks and computes the tuning vectors of the multi-domain norm-referenced encoding (MD-NRE). The role of the FER pathway is to construct a shape- and texture-invariant facial expression landmark detector. The MD-NRE read-out layer encodes the landmark position relative to the selected reference vector and projects it onto the corresponding tuning vector $\bf n_i$.}
	\label{fig:fig2}
\end{figure*}

Figure \ref{fig:fig2} illustrates the developed model architecture. The model consists of three major components: first, a standard convolutional neural network (CNN) for the extraction of image features. Second, a module that selects task-relevant features and computes the difference vectors between the input image and the corresponding norm stimulus. Finally, our proposed high-level readout layer.

A given input image $\bf x$ is fed through the CNN to obtain a representation in terms of hidden units of the network. We reduce the dimension of the representations by applying a \textit{Network Dissection} technique \cite{bau2017network}. The output of the dimensionality reduction is then used as input for the landmark detector. The challenge for the landmark detector is to provide position shifts that are similar for expressions presented on different head shapes. We solved this problem using the two-stream pathway shown in Figure \ref{fig:fig2}. The FR pathway enables the relative encoding of the absolute landmark positions estimated by the FER pathway. The output of the landmark detector is a set of positions of landmarks $l$. This set of vectors constitutes the latent space in which the MD-NRE read-out operates. Further information on implementation and training of the feature extraction, feature reduction and landmark detection are provided in the supplementary materials.

\subsection{Norm-Referenced Encoding and Classifier}
\label{ss:classifier}
The proposed classifier is inspired by results on face-selective neurons in the visual cortex \cite{giese2005physiologically, leopold2006norm}. We assume that the norm vector $\bf r$ has been learned for the individual types of head shapes. The activity $v_m$ of the detector neuron for expression $m$, summing over the different landmarks $l$, is given by the function

\begin{equation}
\label{eq:normrefneruon}
  v_m = \sum_l \left[{\bf d}_l^T {\bf n}_{l, m} \right]_+.
\end{equation}

Here $\textbf{d}_l= \textbf{s}_l - \textbf{r}_l$ is the two-dimensional difference vector for the $l$-th face fragment. The term in the bracket measures its similarity to the learned direction-tuning unit vector ${\bf n}_{l, m}$. The output $v_m$ then depends linearly on the length of the difference vector, which varies monotonically with expression strength. This type of tuning matches the tuning function of neurons in the IT cortex for identity coding \cite{leopold2006norm}. It turns out that a summation of the tuning function outputs over the different fragments provided the most robust results. We obtain the final classification result by determining the norm-referenced neural unit with the maximum output $\hat{m} = {\arg} \max_m v_m$.


\section{Results}
\label{sec:Results}

In the following, we focus on the multi-domain norm-referenced encoding (MD-NRE) and regard its input as a general facial landmark detector with two key characteristics. First, we use 10 landmarks for the construction of the reference and the difference vectors. Second, we use only three landmarks to determine the type, position, and size of the face. These parameters are used to construct the reference vector for the appropriate head shape. Further details about this architecture and its training can be found in the supplementary materials. 

The training of this architecture requires two steps: a first one that trains the reference vectors for all head shapes, and a second one that determines the tuning vectors $\bf n_i$. This training can be accomplished with one image per category (\textit{e.g.}, happy, surprised, etc.) to train each tuning vector, and one image per domain (\textit{e.g.}, human, monkey) to train each reference vector.

We first employ our own dataset to demonstrate \textit{domain generalization} of facial expression recognition (FER) between different head shapes. Specifically, the proposed approach accomplishes high performance in transfer learning, training the system on facial expressions from one head shape only, and testing on other head shapes. We then test our method on a publicly available, larger dataset to investigate its data efficiency.

We also performed ablations studies to investigate the robustness of MD-NRE mechanism against occlusion, and the use of human facial landmark detectors. Further, we compared our MD-NRE model to the NRE without built-in domain transfer capabilities. Results shows that our architecture shows robustness to occlusion and sub-optimal latent spaces. Further, the inclusion of multiple domain references is key for good performance. Details can be found in the supplementary material.
 
\subsection{BFS Dataset}
\label{ss:ImageDataSet}
A valid question is whether the chosen input levels derived from a VGG19 architecture enable robust and precise tracking of facial landmarks across all head shapes. CNN architectures are known to be biased toward textures \cite{geirhos2018generalisation, baker2020local}. Our choice to use the network dissection technique \cite{bau2017network} (see supplementary material) to filter non-face-relevant feature maps does not necessarily remove texture bias. It is unclear a priori whether the remaining feature maps from our read-out layer yield reliable activity patterns for our landmark detectors.

Therefore, we decided to create our own dataset using computer graphics to match facial expressions across different basic head shapes (contribution 3). We selected a human, a cartoon, and a monkey avatar as three basic head shapes and called it the BFS (Basic Face Shapes) dataset. The main goal of this dataset is to investigate the architecture design of our model. Therefore, the specificity of the BFS dataset is to ensure that the displacement of the landmarks remains invariant across all the different head shapes and textures. In total we created 15 identities (5 identities per basic face shape) displaying 7 expressions (see Figure \ref{fig:BFS_example}), resulting in a total of 105 images. All face shapes were animated using the same facial expressions. Correspondence between the meshes of the different avatar types was established by interpolating the meshes of the monkey and the cartoon avatar to match exactly the polygonal structure of the human mesh. Once this correspondence is established, it is straightforward to animate the avatars with the same expressions. (See the supplementary materials for more details.)

\subsection{Domain Generalisation}
\label{ss:transferLearning}
Using our BFS dataset \ref{ss:ImageDataSet}, we test the MD-NRE for \textit{domain generalization}. In this test, we learn facial expressions from a source domain (human) and transfer the tuning vectors to the other domains by updating the reference frames. Here, domains refer to each basic head shape of the BFS dataset (human, monkey, and cartoon). As such, our model has never seen any expressions from the target domains. Nonetheless, the model has knowledge that the target domains exist from the training of the reference frame.

To conduct our test, we select one image of each expression displayed on the human avatar (blue rectangle from Figure \ref{fig:BFS_example}). We use three images, one per domain, to train the reference vectors, and 7 images, one per category, to train the tuning vectors. However, as the neutral expression on the human identity is already selected for the training of the human reference vector, we effectively add only 6 images to the training set. Consequently, we train the classifier using a total of 9 images. 

Nonetheless, it is fair to wonder if our BFS cross-domain dataset is too similar to human head shapes. The question emerges whether state-of-the-art FER models can accurately classify our non-human head shapes. Therefore, we test the performance of multiple available FER architectures and trained a CORNet-S model for comparisons (as it is argued to be a closer brain-like architecture) \cite{kubilius2018cornet}. (See the supplementary materials for implementation and training details.)

\begin{table}[t]
	\begin{tabular}{l r r r r}
		\hline\noalign{\smallskip}
		Model                                   & Type & Dataset    & \#im  & Acc\\
		\hline
        SVM-LBP \cite{luo2013facial}            & SVM & BFS         & 9     & 14.3 \\
        ResNet50v2 \cite{ngo2020facial}         & CNN & Affectnet   & 280k+ & 52.4 \\
        CORNet-S                                & CNN & Affectnet   & 280k+ & 63.1 \\
        DAN \cite{wen2021dan}                   & CNN & AffectNet   & 440k  & 37.4 \\
        EfficientFace \cite{Zhao2021RobustLF}   & CNN & AffectNet   & 440k  & 33.8 \\
		\rule{0pt}{2.5ex}MD-NRE                 &     & BFS         & 9     & \textbf{100}\\
  		\hline
	\end{tabular}
 	\caption{Classification accuracies (\%) for our MD-NRE model and different standard CNN architectures.
   Letters refer to the training condition with: B (only BFS), A (only AffectNet) and C (AffectNet + BFS)}
	\label{table:BFS_results}
\end{table}

We obtain perfect classification accuracy (100\%) for the test dataset. This implies a perfect transfer of the learned expressions from one human identity to all the other head shapes, without a need for training of the expressions on the non-human head shapes (Table \ref{table:BFS_results}). This result proves our assumption that facial expression tuning vectors can be transferred across different head shapes using only a single neutral face shape within an appropriate feature space. It also proves that our model reliably tracks the position of facial landmarks independently of face shapes and textures.

Table \ref{table:BFS_results} shows that none of the other models performed as well as ours on the transfer learning task. The CORNet-S model reaches the best accuracy with a limited performance of 63.1\%, and the SVM model performs the worst, only reaching chance level classification performance (14.3\%). See the supplementary materials for a detailed discussion of this poor result.

\subsection{Analogous Encoding}
\label{ss:RobustLevel}
The second test aims at assessing the second assumption of the model. Contrary to the output layers of typical DNN classification models, which return estimates for class probabilities (softmax readout), face-selective cortical neurons show an almost linear dependence of activity on the distance of the face from the average face \cite{leopold2006norm, koyano2021dynamic, freiwald2021neuroscience}. Likewise, in our architecture, the output variables $v_m$ depend linearly on the length of the difference vectors {\bf d}, which covary with expression strength. Such \textit{analogous encoding} is especially interesting for social interaction, where even small and subtle facial movements might be used to perceive a person's intent. This principle might underlie our ability to perceive complex and subtle facial movements such as a smirk. Furthermore, encoding of expression strength seems useful for different technical applications. One example would be a human closed-loop interaction system, such as using facial expression recognition data to drive computer animations, where expression strength is often modeled by a superposition of action units or blendshapes. In addition, expression strength influences the reliability of expression categorization: weak expressions are more difficult to classify than extreme expressions.

We test the linearity of the encoding by leveraging our BFS dataset to simulate expression strength for each basic head shape. For this purpose, we extended the BFS dataset by adding versions of all expressions with three extra levels (25\%, 50\%, and 75\%) of expression strength. These intermediate levels were created by blending (linearly scaling the displacements between the corresponding mesh polygons between the neutral and the 100 \% expressive face for the same head shape). (See the supplementary materials for an illustration.) The resulting dataset (BFS-L) has a total of 75 testing images. We use the same 9 training images as in the previous validation experiments. 

\begin{table}[t]
	\centering
	\begin{tabular}{l r r r r}
		\hline\noalign{\smallskip}
		Model       & BFS-L         & Human         & Monkey      & Cartoon\\
		\hline
       CORnet-S    & 46            & 62.5          & 44          & 32\\
		MD-NRE      & \textbf{78}   & \textbf{92}   & \textbf{68} & \textbf{76} \\
  		\hline
	\end{tabular}
 	\caption{Classification accuracy (\%) for our MD-NRE and the CORnetS-A on the expression strength level test (BFS-L) for each head shapes.}
	\label{table:BFS_expression_strenght_results}
\end{table}

\begin{figure}[h]
	\centering
	\includegraphics[width=.6\linewidth]{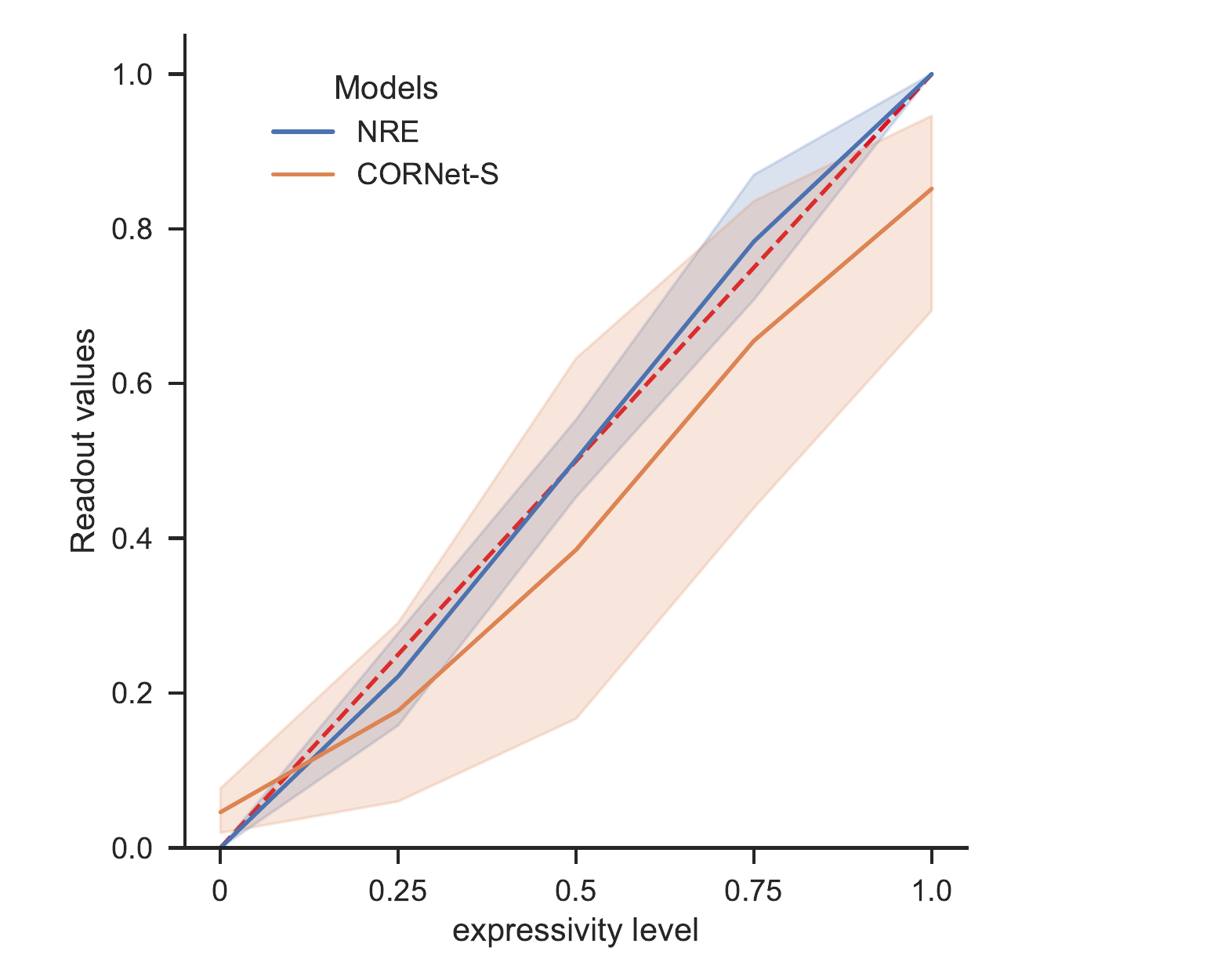}
	\caption{Readout values of the multi-domain norm-referenced encoding (MD-NRE) and the CORnetS-A model as a function of the expression strength level for each expression on the Human Avatar. The red dashed line corresponds to perfect linearity.}
	\label{fig:expression_strengh}
\end{figure}

Figure \ref{fig:expression_strengh} compares the readout activity values of our MD-NRE and the CORNetS models (the best CNN model from the previous experiment). While both models show a monotonic increase of the output variables with the expression strength, our model shows greater linearity. Moreover, Table \Ref{table:BFS_expression_strenght_results} shows that the classification accuracy (collapsed over all tested expression strength levels) of our model outperforms the CORNetS architecture, indicating higher robustness to variations of expression strength. Nonetheless, as expected, the overall performance for the MD-NRE model is reduced compared to using only the 100\% expression strength test images (Table \ref{table:BFS_results}). This accuracy reduction is mainly caused by a reduced performance on the weaker expression strength level images (see supplementary materials). 

\subsection{Efficient Learning}
\label{ss:dataEfficiency}
The third test aims to scale up the model and to conduct a first benchmark test of our approach with the literature. We select the FERG \cite{aneja2016modeling} dataset which consists of stylized comic characters with annotated facial expressions. The FERG dataset is suitable for our simple landmark detector as the pose and the illumination are fixed. The dataset contains 55,767 images across 6 different head shapes and 7 expressions. The dataset is split into a training set (43,767), a validation set (6k), and a test set (6K). 

To apply our MD-NRE, we train a reference vector for each avatar identity and test our model's learning efficiency. This task is thus an assessment of the benefit of using reference frames as inductive bias. Nonetheless, the variance of the head shapes among the avatars of the FERG dataset is arguably higher than among human identities. Therefore, we also test our method to transfer facial expressions from a single avatar head to the others. 

We use 12 images to train our models. One neutral pose for each avatar type (6 images) to train the reference vectors, and one for each expression (7 images) to train the tuning vectors. As before, the neutral expression is already included in the set used to train the reference vectors.

We test the efficiency of our model using two approaches. First, we train our model as a standard FER approach. Therefore, the tuning vectors are trained independently of the avatar type (referred to as MD-NRE-I). Here the model can select any expression that might help classification accuracy. Second, we train our classifier using a transfer learning task by using only a single avatar to train the tuning vectors, referred to by MD-NRE-$<$avatar\_name$>$. Here, the models learn facial expressions from the subset of images belonging to the selected avatar. 

In addition, we select the training expressions in two ways: In one condition, indicated by {\em First}, we use the first picture of the relevant expression class for training. This has the disadvantage that this first picture might not be the most typical one for that expression category. To remove this limitation, we perform a second set of tests where we choose for each class an optimized template stimulus (indicated by {\em Optimized}). This stimulus is determined by iterating through the training dataset and retaining the class-specific training picture that results in the highest training accuracy after a single training epoch.

\begin{table}[t]
	\centering
    \resizebox{0.475\textwidth}{!}{%
	\begin{tabular}{l r r r r r}
		\hline\noalign{\smallskip}
		Models & \multicolumn{2}{c}{First}  & \multicolumn{2}{c}{Optimized}         & \# im\\
                                            & Train & Test  & Train & Test          &   \\
		\hline
        DeepExpr \cite{aneja2016modeling}   & -     & -     & -     & 89            & 43k+\\
        ACNN \cite{minaee2021deep}          & -     & -     & -     & 99.3          & 43k+\\
        VGAN \cite{chen2018vgan}            & -     & -     & -     & 100           & 43k+ \\
        \rule{0pt}{2.5ex}MD-NRE-I           & 73.1  & 73.1  & 92.3  & \textbf{92.2} & \textbf{12}\\
        \rule{0pt}{2.5ex}MD-NRE-Aia         & 76.7  & 76.4  & 84.6  & 84.2          & 12\\
        MD-NRE-Bonnie                       & 70.4  & 70.4  & 85    & 84.4          & 12\\
        MD-NRE-Jules                        & 80.6  & 79.9  & 80.5  & 79.7          & 12\\
        MD-NRE-Malcolm                      & 73.3  & 73.3  & 80.4  & 80            & 12\\
        MD-NRE-Mery                         & 71.6  & 71,1  & 81.9  & 82.2          & 12\\
        MD-NRE-Ray                          & 66.6  & 66.5  & 71.9  & 71.6          & 12\\
        \hline
	\end{tabular}}
 	\caption{Classification accuracy (\%) for our MD-NRE models on the FERG dataset and the reported test classification accuracy from different FER models trained on the FERG dataset}
	\label{table:FERG_results}
\end{table}

Table \ref{table:FERG_results} shows the accuracy of the classifications for the different training schemes. Using the first images in the dataset for each class (condition {\em First}), we obtain a testing accuracy of 73\%. This was in the range of the accuracies for training with individual avatars, which range from 66\% to 79\%. While these results are not state-of-the-art (100\% reported by \cite{chen2018vgan}), this test proves that our model can cope with a larger dataset. When the model have access to every domain (MD-NRE-I), the test accuracy reaches a value of 92.15\%, which exceeds the accuracy (89.02\%) reported in the original paper on the FERG dataset \cite{aneja2016modeling}. Moreover, the best transfer learning model (MD-NRE-Bonnie) reaches 84.42\% accuracy. This is an encouraging transfer learning result for a classifier trained with only 12 images, and where all the expressions are from a single avatar type.


\section{Conclusion}
We present a proof-of-concept study that tests the use of a biologically inspired principle: norm-referenced encoding (NRE) for multi-domain transfer in facial expression classification. There is strong evidence for such representations in the brain (\textit{e.g.} \cite{leopold2006norm, rhodes2011adaptive, koyano2021dynamic, caggiano2016mirror, mattar2016varying}. We hypothesize that this encoding principle explains how the human brain accomplishes high data-efficiency in transfer learning. We validate our hypothesis by illustrating how our model recognizes facial expressions on novel head shapes, an easy task for humans.

We demonstrate that this encoding principle can be implemented in the context of a deep network architecture that consists of two streams, one that identifies the type of the head shapes and the corresponding reference vectors, and a second one that computes the difference vectors and the activity of the NRE encoding neurons. The input of these two pathways is given by a landmark detector.

We also demonstrate its linear dependence on deviation from the reference frames makes this encoding especially interesting for tasks requiring a strong estimate of confidence, \textit{e.g.}, human social interactions, and medical applications. Facial expression perception is a prime example: even subtle facial cues might reveal information about emotion. And in many social situations, the expressiveness about the displayed emotion has important behavioral consequences (e.g faked vs. real smiles, etc.).

A major goal of our work is to verify if a principle that seems relevant in neuroscience might be helpful in computer vision and machine learning, especially for cases in which large amounts of training data are not available.

We hope that this work demonstrates the value of biologically inspired methods in machine-learning generalization across different domains. We acknowledge that substantial additional work is required to make these methods applicable to a wider spectrum and more complex machine-learning problems. Nonetheless, this work is a step towards a more generic end-to-end machine learning model exploiting NRE. 

\section*{Acknowledgments}
We would like to thank Michaël Morret, Marie Stettler, Luigi Gresele, Julius von Kügelgen and Jesse St. Amand for their general advice and feedback on the paper.

This was was supported in part by the European Research Council ERC 2019-SYG under EU Horizon 2020 research and innovation programme (grant agreement No. 856495, RELEVANCE). Martin Giese was also supported by BMBF FKZ 01GQ1704 and BMG: SSTeP-KiZ (grant no. ZMWI1- 2520DAT700). The authors thank the International Max Planck Research School for Intelligent Systems (IMPRS-IS) for supporting Michael Stettler and Alexander Lappe. Nvidia Corp.

\bibliographystyle{unsrt}  
\bibliography{references}

\begin{thebibliography}{10}

\bibitem{wilson2012social}
Edward~O Wilson et~al.
\newblock {\em The social conquest of earth}.
\newblock WW Norton \& Company, 2012.

\bibitem{zhu2013dissimilar}
Qi~Zhu, Koen Nelissen, Jan Van~den Stock, Fran{\c{c}}ois-Laurent De~Winter,
  Karl Pauwels, Beatrice de~Gelder, Wim Vanduffel, and Mathieu Vandenbulcke.
\newblock Dissimilar processing of emotional facial expressions in human and
  monkey temporal cortex.
\newblock {\em Neuroimage}, 66:402--411, 2013.

\bibitem{taubert2021shape}
Nick Taubert, Michael Stettler, Ramona Siebert, Silvia Spadacenta, Louisa
  Sting, Peter Dicke, Peter Thier, and Martin~A Giese.
\newblock Shape-invariant encoding of dynamic primate facial expressions in
  human perception.
\newblock {\em Elife}, 10:e61197, 2021.

\bibitem{ekman1976measuring}
Paul Ekman and Wallace~V Friesen.
\newblock Measuring facial movement.
\newblock {\em Environmental psychology and nonverbal behavior}, 1:56--75,
  1976.

\bibitem{jack2016four}
Rachael~E Jack, Wei Sun, Ioannis Delis, Oliver~GB Garrod, and Philippe~G
  Schyns.
\newblock Four not six: Revealing culturally common facial expressions of
  emotion.
\newblock {\em Journal of Experimental Psychology: General}, 145(6):708, 2016.

\bibitem{rhodes1987identification}
Gillian Rhodes, Susan Brennan, and Susan Carey.
\newblock Identification and ratings of caricatures: Implications for mental
  representations of faces.
\newblock {\em Cognitive psychology}, 19(4):473--497, 1987.

\bibitem{gao2003facial}
Yongsheng Gao, Maylor~KH Leung, Siu~Cheung Hui, and Mario~W Tananda.
\newblock Facial expression recognition from line-based caricatures.
\newblock {\em IEEE Transactions on Systems, Man, and Cybernetics-Part A:
  Systems and Humans}, 33(3):407--412, 2003.

\bibitem{kaye2017emojis}
Linda~K Kaye, Stephanie~A Malone, and Helen~J Wall.
\newblock Emojis: Insights, affordances, and possibilities for psychological
  science.
\newblock {\em Trends in cognitive sciences}, 21(2):66--68, 2017.

\bibitem{zhao2019event}
Jiayin Zhao, Qi~Meng, Licong An, and Yifang Wang.
\newblock An event-related potential comparison of facial expression processing
  between cartoon and real faces.
\newblock {\em PLoS One}, 14(1):e0198868, 2019.

\bibitem{zhang2021influence}
Shu Zhang, Xinge Liu, Xuan Yang, Yezhi Shu, Niqi Liu, Dan Zhang, and Yong-Jin
  Liu.
\newblock The influence of key facial features on recognition of emotion in
  cartoon faces.
\newblock {\em Frontiers in psychology}, 12:687974, 2021.

\bibitem{wardle2020rapid}
Susan~G Wardle, Jessica Taubert, Lina Teichmann, and Chris~I Baker.
\newblock Rapid and dynamic processing of face pareidolia in the human brain.
\newblock {\em Nature communications}, 11(1):4518, 2020.

\bibitem{geirhos2018generalisation}
Robert Geirhos, Carlos R~Medina Temme, Jonas Rauber, Heiko~H Sch{\"u}tt,
  Matthias Bethge, and Felix~A Wichmann.
\newblock Generalisation in humans and deep neural networks.
\newblock {\em arXiv preprint arXiv:1808.08750}, 2018.

\bibitem{baker2020local}
Nicholas Baker, Hongjing Lu, Gennady Erlikhman, and Philip~J Kellman.
\newblock Local features and global shape information in object classification
  by deep convolutional neural networks.
\newblock {\em Vision research}, 172:46--61, 2020.

\bibitem{dosovitskiy2020image}
Alexey Dosovitskiy, Lucas Beyer, Alexander Kolesnikov, Dirk Weissenborn,
  Xiaohua Zhai, Thomas Unterthiner, Mostafa Dehghani, Matthias Minderer, Georg
  Heigold, Sylvain Gelly, et~al.
\newblock An image is worth 16x16 words: Transformers for image recognition at
  scale.
\newblock {\em arXiv preprint arXiv:2010.11929}, 2020.

\bibitem{rhodes2011adaptive}
Gillian Rhodes and David~A Leopold.
\newblock Adaptive norm-based coding of face identity.
\newblock {\em The Oxford handbook of face perception}, pages 263--286, 2011.

\bibitem{leopold2006norm}
David~A Leopold, Igor~V Bondar, and Martin~A Giese.
\newblock Norm-based face encoding by single neurons in the monkey
  inferotemporal cortex.
\newblock {\em Nature}, 442(7102):572--575, 2006.

\bibitem{koyano2021dynamic}
Kenji~W Koyano, Adam~P Jones, David~BT McMahon, Elena~N Waidmann, Brian~E Russ,
  and David~A Leopold.
\newblock Dynamic suppression of average facial structure shapes neural tuning
  in three macaque face patches.
\newblock {\em Current Biology}, 31(1):1--12, 2021.

\bibitem{giese2005physiologically}
Martin~A Giese and David~A Leopold.
\newblock Physiologically inspired neural model for the encoding of face
  spaces.
\newblock {\em Neurocomputing}, 65:93--101, 2005.

\bibitem{chang2017code}
Le~Chang and Doris~Y Tsao.
\newblock The code for facial identity in the primate brain.
\newblock {\em Cell}, 169(6):1013--1028, 2017.

\bibitem{freiwald2021neuroscience}
Winrich~A Freiwald and Haruo Hosoya.
\newblock Neuroscience: A face’s journey through space and time.
\newblock {\em Current Biology}, 31(1):R13--R15, 2021.

\bibitem{stettler2020physiologically}
Michael Stettler, Nick Taubert, Tahereh Azizpour, Ramona Siebert, Silvia
  Spadacenta, Peter Dicke, Peter Thier, and Martin~A Giese.
\newblock Physiologically-inspired neural circuits for the recognition of
  dynamic faces.
\newblock In {\em International Conference on Artificial Neural Networks},
  pages 168--179. Springer, 2020.

\bibitem{mollahosseini2017affectnet}
Ali Mollahosseini, Behzad Hasani, and Mohammad~H Mahoor.
\newblock Affectnet: A database for facial expression, valence, and arousal
  computing in the wild.
\newblock {\em IEEE Transactions on Affective Computing}, 10(1):18--31, 2017.

\bibitem{courville2013fer}
PLC Courville, A~Goodfellow, IJM Mirza, and Y~Bengio.
\newblock Fer-2013 face database.
\newblock {\em Universit de Montreal: Montr{\'e}al, QC, Canada}, 2013.

\bibitem{lucey2010extended}
Patrick Lucey, Jeffrey~F Cohn, Takeo Kanade, Jason Saragih, Zara Ambadar, and
  Iain Matthews.
\newblock The extended cohn-kanade dataset (ck+): A complete dataset for action
  unit and emotion-specified expression.
\newblock In {\em 2010 ieee computer society conference on computer vision and
  pattern recognition-workshops}, pages 94--101. IEEE, 2010.

\bibitem{calvo2008facial}
Manuel~G Calvo and Daniel Lundqvist.
\newblock Facial expressions of emotion (kdef): Identification under different
  display-duration conditions.
\newblock {\em Behavior research methods}, 40(1):109--115, 2008.

\bibitem{aneja2016modeling}
Deepali Aneja, Alex Colburn, Gary Faigin, Linda Shapiro, and Barbara Mones.
\newblock Modeling stylized character expressions via deep learning.
\newblock In {\em Asian conference on computer vision}, pages 136--153.
  Springer, 2016.

\bibitem{ionescu2013local}
Radu~Tudor Ionescu, Marius Popescu, and Cristian Grozea.
\newblock Local learning to improve bag of visual words model for facial
  expression recognition.
\newblock In {\em Workshop on challenges in representation learning, ICML}.
  Citeseer, 2013.

\bibitem{georgescu2019local}
Mariana-Iuliana Georgescu, Radu~Tudor Ionescu, and Marius Popescu.
\newblock Local learning with deep and handcrafted features for facial
  expression recognition.
\newblock {\em IEEE Access}, 7:64827--64836, 2019.

\bibitem{giannopoulos2018deep}
Panagiotis Giannopoulos, Isidoros Perikos, and Ioannis Hatzilygeroudis.
\newblock Deep learning approaches for facial emotion recognition: A case study
  on fer-2013.
\newblock {\em Advances in Hybridization of Intelligent Methods: Models,
  Systems and Applications}, pages 1--16, 2018.

\bibitem{niu2021facial}
Ben Niu, Zhenxing Gao, and Bingbing Guo.
\newblock Facial expression recognition with lbp and orb features.
\newblock {\em Computational Intelligence and Neuroscience}, 2021:1--10, 2021.

\bibitem{ashir2020facial}
Abubakar~M Ashir, Alaa Eleyan, and Bayram Akdemir.
\newblock Facial expression recognition with dynamic cascaded classifier.
\newblock {\em Neural Computing and Applications}, 32:6295--6309, 2020.

\bibitem{zhao2018transfer}
Hang Zhao, Qing Liu, and Yun Yang.
\newblock Transfer learning with ensemble of multiple feature representations.
\newblock In {\em 2018 IEEE 16th international conference on software
  engineering research, management and applications (SERA)}, pages 54--61.
  IEEE, 2018.

\bibitem{wang2018unsupervised}
Xiaoqing Wang, Xiangjun Wang, and Yubo Ni.
\newblock Unsupervised domain adaptation for facial expression recognition
  using generative adversarial networks.
\newblock {\em Computational intelligence and neuroscience}, 2018, 2018.

\bibitem{ribera2017facial}
Roger Blanco~I Ribera, Eduard Zell, John~P Lewis, Junyong Noh, and Mario
  Botsch.
\newblock Facial retargeting with automatic range of motion alignment.
\newblock {\em ACM Transactions on graphics (TOG)}, 36(4):1--12, 2017.

\bibitem{chaudhuri2019joint}
Bindita Chaudhuri, Noranart Vesdapunt, and Baoyuan Wang.
\newblock Joint face detection and facial motion retargeting for multiple
  faces.
\newblock In {\em Proceedings of the IEEE/CVF Conference on Computer Vision and
  Pattern Recognition}, pages 9719--9728, 2019.

\bibitem{kouw2019review}
Wouter~M Kouw and Marco Loog.
\newblock A review of domain adaptation without target labels.
\newblock {\em IEEE transactions on pattern analysis and machine intelligence},
  43(3):766--785, 2019.

\bibitem{wang2022generalizing}
Jindong Wang, Cuiling Lan, Chang Liu, Yidong Ouyang, Tao Qin, Wang Lu, Yiqiang
  Chen, Wenjun Zeng, and Philip Yu.
\newblock Generalizing to unseen domains: A survey on domain generalization.
\newblock {\em IEEE Transactions on Knowledge and Data Engineering}, 2022.

\bibitem{wang2020generalizing}
Yaqing Wang, Quanming Yao, James~T Kwok, and Lionel~M Ni.
\newblock Generalizing from a few examples: A survey on few-shot learning.
\newblock {\em ACM computing surveys (csur)}, 53(3):1--34, 2020.

\bibitem{vetter1995separate}
Thomas Vetter and Nikolaus~F Troje.
\newblock A separate linear shape and texture space for modeling
  two-dimensional images of human faces.
\newblock 1995.

\bibitem{vetter1997bootstrapping}
Thomas Vetter, Michael~J Jones, and Tomaso Poggio.
\newblock A bootstrapping algorithm for learning linear models of object
  classes.
\newblock In {\em Proceedings of IEEE Computer Society Conference on Computer
  Vision and Pattern Recognition}, pages 40--46. IEEE, 1997.

\bibitem{blanz1999morphable}
Volker Blanz and Thomas Vetter.
\newblock A morphable model for the synthesis of 3d faces.
\newblock In {\em Proceedings of the 26th annual conference on Computer
  graphics and interactive techniques}, pages 187--194, 1999.

\bibitem{li2017learning}
Tianye Li, Timo Bolkart, Michael~J Black, Hao Li, and Javier Romero.
\newblock Learning a model of facial shape and expression from 4d scans.
\newblock {\em ACM Trans. Graph.}, 36(6):194--1, 2017.

\bibitem{egger20203d}
Bernhard Egger, William~AP Smith, Ayush Tewari, Stefanie Wuhrer, Michael
  Zollhoefer, Thabo Beeler, Florian Bernard, Timo Bolkart, Adam Kortylewski,
  Sami Romdhani, et~al.
\newblock 3d morphable face models—past, present, and future.
\newblock {\em ACM Transactions on Graphics (TOG)}, 39(5):1--38, 2020.

\bibitem{bulat2017far}
Adrian Bulat and Georgios Tzimiropoulos.
\newblock How far are we from solving the 2d \& 3d face alignment problem? (and
  a dataset of 230,000 3d facial landmarks).
\newblock In {\em International Conference on Computer Vision}, 2017.

\bibitem{kartynnik2019real}
Yury Kartynnik, Artsiom Ablavatski, Ivan Grishchenko, and Matthias Grundmann.
\newblock Real-time facial surface geometry from monocular video on mobile
  gpus.
\newblock {\em arXiv preprint arXiv:1907.06724}, 2019.

\bibitem{bau2017network}
David Bau, Bolei Zhou, Aditya Khosla, Aude Oliva, and Antonio Torralba.
\newblock Network dissection: Quantifying interpretability of deep visual
  representations.
\newblock In {\em Proceedings of the IEEE conference on computer vision and
  pattern recognition}, pages 6541--6549, 2017.

\bibitem{kubilius2018cornet}
J~Kubilius, M~Schrimpf, A~Nayebi, D~Bear, DLK Yamins, and JJ~DiCarlo.
\newblock Cornet: Modeling the neural mechanisms of core object recognition.
  biorxiv, 408385, 2018.

\bibitem{luo2013facial}
Yuan Luo, Cai-ming Wu, and Yi~Zhang.
\newblock Facial expression recognition based on fusion feature of pca and lbp
  with svm.
\newblock {\em Optik-International Journal for Light and Electron Optics},
  124(17):2767--2770, 2013.

\bibitem{ngo2020facial}
Quan~T Ngo and Seokhoon Yoon.
\newblock Facial expression recognition based on weighted-cluster loss and deep
  transfer learning using a highly imbalanced dataset.
\newblock {\em Sensors}, 20(9):2639, 2020.

\bibitem{wen2021dan}
Zhengyao Wen, Wenzhong Lin, Tao Wang, and Ge~Xu.
\newblock Distract your attention: Multi-head cross attention network for
  facial expression recognition.
\newblock {\em CoRR}, abs/2109.07270, 2021.

\bibitem{Zhao2021RobustLF}
Zengqun Zhao, Qingshan Liu, and Feng Zhou.
\newblock Robust lightweight facial expression recognition network with label
  distribution training.
\newblock In {\em AAAI Conference on Artificial Intelligence}, 2021.

\bibitem{minaee2021deep}
Shervin Minaee, Mehdi Minaei, and Amirali Abdolrashidi.
\newblock Deep-emotion: Facial expression recognition using attentional
  convolutional network.
\newblock {\em Sensors}, 21(9):3046, 2021.

\bibitem{chen2018vgan}
Jiawei Chen, Janusz Konrad, and Prakash Ishwar.
\newblock Vgan-based image representation learning for privacy-preserving
  facial expression recognition.
\newblock In {\em Proceedings of the IEEE conference on computer vision and
  pattern recognition workshops}, pages 1570--1579, 2018.

\bibitem{caggiano2016mirror}
Vittorio Caggiano, Falk Fleischer, Joern~K Pomper, Martin~A Giese, and Peter
  Thier.
\newblock Mirror neurons in monkey premotor area f5 show tuning for critical
  features of visual causality perception.
\newblock {\em Current Biology}, 26(22):3077--3082, 2016.

\bibitem{mattar2016varying}
Marcelo~G Mattar, David~A Kahn, Sharon~L Thompson-Schill, and Geoffrey~K
  Aguirre.
\newblock Varying timescales of stimulus integration unite neural adaptation
  and prototype formation.
\newblock {\em Current Biology}, 26(13):1669--1676, 2016.

\end{thebibliography}

\end{document}